\documentclass[letterpaper, 10 pt, journal, twoside]{IEEEtran}
\ifCLASSINFOpdf
\else
\fi

\usepackage{hyperref} 
\usepackage{graphicx}
\usepackage{multirow}
\usepackage{amssymb}
\usepackage{pifont}
\usepackage{xcolor}
\newcommand{\xmark}{\ding{55}}
\usepackage{amsmath}
\usepackage{adjustbox}

\begin{document}
%
\title{Learning Context-Aware Neural ODE Dynamics for Adaptive Robotic Control}
%
%
%

\author{Shao-Yi Yu$^{1*}$, Jen-Wei Wang$^{1*}$, Maya Horii$^{1}$, Masayoshi Tomizuka$^{1}$ and Vikas Garg$^{2}$%
\thanks{Manuscript received: February, 1, 2026; Revised April, 12, 2026; Accepted May, 11, 2026.}
\thanks{ This paper was recommended for publication by
Editor Jens Kober upon evaluation of the Associate Editor and Reviewers
comments.} 
\thanks{$^{*}$Shao-Yi Yu and Jen-Wei Wang are co-first authors.}
\thanks{Emails: \texttt{\{syyu410, jwwang\}@berkeley.edu}}
\thanks{$^{1}$Authors are with Department of Mechanical Engineering, University of California, Berkeley, CA, USA}%
\thanks{$^{2} $Author is with Aalto University and YaiYai Ltd, Finland}
\thanks{Digital Object Identifier (DOI): see top of this page.}
}
%
%

\markboth{IEEE Robotics and Automation Letters. Preprint Version. Accepted May, 2026}
{Yu \MakeLowercase{\textit{et al.}}: Learning Context-Aware Neural ODE Dynamics for
Adaptive Robotic Control} 

%



\maketitle

\begin{abstract}
Robotic systems deployed in uncertain and dynamically changing environments often face variations in contact conditions, aerodynamic effects, and external disturbances that challenge reliable control. To remain effective under model-based control, these systems require dynamics models that can adapt to such changes, especially when direct access to complete environmental information is limited. To enable adaptability and facilitate integration with model predictive control, we propose a context-aware dynamics model based on neural ordinary differential equations, which infers environmental factors from state–action histories using a two-phase training procedure. We validate the approach across diverse robotic platforms, including a quadrotor in simulation, as well as a Sphero BOLT robot and a Fanuc manipulator in real-world experiments. The results demonstrate that our method effectively adapts to temporally and spatially varying environmental changes across different tasks. Videos are available \href{https://youtu.be/PY0sNyF2rqE}{here}, and the source code is available at our \href{https://github.com/syyu410-yu/context-aware-neural-ode-control.git}{GitHub repository}.
\end{abstract}

\begin{IEEEkeywords}
Machine Learning for Robot Control, Reinforcement Learning, Adaptive Control, Learning-based MPC
\end{IEEEkeywords}

%
\IEEEpeerreviewmaketitle

\section{Introduction}
%
%
%
%
\IEEEPARstart{R}{obots} are often deployed in uncertain and dynamically changing environments. For example, mobile robots, such as ground vehicles or drones, often need to adapt to varying terrain conditions and wind disturbances, while manipulators may experience changes in payload when transporting items in warehouses. These environmental variations make real-time adaptability essential. However, achieving such adaptability remains challenging for model-based controllers, as they rely on accurate dynamics models for planning over long horizons \cite{seo2020trajectory, nagabandi2018learning}. Furthermore, many environmental variations cannot be fully detected using onboard sensors, making it important for the system to infer hidden environmental factors from limited data. 

Previous efforts in in-context reinforcement learning (RL) have led to major advances in adapting to different environments based on past trajectories \cite{liang2023context, zhang2025dynamics, meta_baseline}. A line of research in adaptive model-free RL proposes specially designed adaptive modules, known as Rapid Motor Adaptation (RMA), to encode environmental information in RL policy \cite{RMA, drone_RMA, in_hand_RMA}. However, model-free RL methods often struggle to explicitly incorporate desired trajectories or constraints, such as collision avoidance, into the policy, which in turn requires large amounts of exploratory data.

Several model-based RL approaches have been proposed \cite{seo2020trajectory, context_aware, modelbased_incontext}; however, they often model dynamics over a predefined discrete time domain, which overlooks the continuously-evolving dynamics of rigid-body robotic systems \cite{hamiltonian}. Since the dynamics of these systems are typically governed by ordinary differential equations (ODE), neural ordinary differential equations (NODE) \cite{NODE}, which learn (first-order) derivatives and compute system states using numerical integrators, are well-suited for modeling continuous-time dynamics. The approach of modeling derivatives has also shown success in time-series prediction tasks across various domains \cite{flowmatching, zhang2024trajectory, cranmer2020lagrangian}. In robotics, learning dynamics with NODE has demonstrated robustness to noisy and irregular data in standard RL tasks \cite{yildiz2021continuous}. However, most NODE-related work either focuses primarily on data efficiency or requires prior system knowledge for successful adaptation. \cite{pmlr-v229-kasaei23a, duong2022adaptivecontrolse3hamiltonian}.

In this work, we propose a context-aware NODE-based dynamics model that learns full dynamics without prior physics knowledge. To improve training efficiency, we incorporate an RMA-style adaptive module that decouples the learning of environmental conditions from system dynamics, thereby reducing the input dimensions required for training the NODE. As a result, our approach enables efficient adaptation with low computational overhead while retaining the advantages of NODE, including smooth dynamics and reduced compounding errors. This makes the context-aware dynamics model well-suited for integration with model predictive control (MPC) to determine optimal actions. We validate the approach across multiple robotic platforms. Experimental results show that our model outperforms baseline methods on a quadrotor flying through varying wind fields in simulation. We also deploy the model on two real-world systems: a Sphero BOLT navigating across different ground textures and a Fanuc manipulator pushing an object with a changing or unseen center of mass (CoM) during operation. In both cases, our method achieves higher accuracy and robustness than its non-adaptive counterpart. The videos are available \href{https://youtu.be/PY0sNyF2rqE}{here}.

%

\section{RELATED WORK}

\textbf{Reliance on Physics-Based Nominal Model.}
Adaptive dynamics modeling for robotic systems has been explored through a variety of approaches aimed at handling changing and partially observed environments. Classical adaptive control \cite{Hanover_2022} and online estimation with known features \cite{9867457} can compensate for disturbances with minimal computational overhead, but their performance is limited by the accuracy of the nominal model and the prior specification of features. Some recent works incorporate NODE variants, such as KNODE \cite{pmlr-v205-jiahao23a} and Hamiltonian NODE \cite{duong2022adaptivecontrolse3hamiltonian}, to model residual dynamics not captured by the nominal model. While more expressive, these approaches still rely on the availability of a known physics model. In contrast, our approach removes this dependency by learning the full system dynamics directly from data, without requiring prior physics knowledge.

\textbf{Test-Time Adaptation vs. Train-Time Adaptation.}
A major distinction in adaptive modeling lies in whether adaptation occurs during deployment or is amortized into training. Several methods perform test-time adaptation by updating model parameters online. For example, meta-learning approaches obtain expressive nonlinear features offline and adapt linear parameters online \cite{richards2021adaptivecontrolorientedmetalearningnonlinearsystems}, while graph-based Koopman embeddings update linear transition matrices via least-squares \cite{module1}. These methods are computationally efficient at deployment but are restricted to updating only a linear layer, leading to poor adaptation under highly nonlinear regime shifts such as contact-rich interactions or rapidly changing terrain. Other approaches adapt model weights online via gradient descent \cite{shi2021metaadaptivenonlinearcontroltheory, nagabandi2018learning, metalearning3}, improving expressiveness but at the cost of higher computational overhead, potential instability from noisy gradients, and increased risk of overfitting to recent observations. Alternatively, train-time adaptation methods encode adaptability into the model during training to avoid parameter updates at deployment. These include architectures such as transformers that leverage in-context learning from state-action history \cite{xiao2024anycaranywherelearninguniversal}, Neural Process context encoders that incorporate terrain-specific and robot-specific context \cite{guttikonda2024contextconditionalnavigationlearningbasedterrain}, and multimodal latent mappers that fuse sensory inputs \cite{10341771}. Additional training objectives, such as joint forward and backward dynamics prediction \cite{context_aware}, further improve representation learning. However, these end-to-end approaches may struggle with learning complex mappings due to entangled system and environmental dynamics, as well as high-dimensional histories inputs.

\textbf{Leveraging Privileged Environmental Information.}
Recent work demonstrates that incorporating privileged environmental information during training can significantly improve the learning of adaptive representations \cite{O_Connell_2022, meta_baseline, xie2023hierarchicalmetalearningbasedadaptivecontroller, levy2025metalearningonlinedynamicsmodel}. By guiding the model to disentangle environment-dependent factors from system dynamics, these methods enable more effective adaptation at deployment. However, many existing approaches do not explicitly integrate this advantage with continuous-time modeling frameworks, limiting their compatibility with certain control strategies.

\textbf{NODE-Based Dynamics Modeling.}
A growing line of research explores the use of NODE for robot modeling and control, including applications in continuous forward kinematics for soft robots \cite{pmlr-v229-kasaei23a} and dynamics learning for manipulation \cite{10802736}. These works highlight the benefits of continuous-time representations and improved data efficiency. However, they primarily focus on stationary settings and do not address adaptation to changing environments. While some adaptive extensions using NODE variants exist \cite{pmlr-v205-jiahao23a, duong2022adaptivecontrolse3hamiltonian}, they typically rely on residual modeling with known physics priors, as discussed earlier. In contrast, our method leverages NODE to learn full system dynamics in an adaptive framework without requiring prior models, while also benefiting from continuous-time structure that integrates naturally with MPC.

\textbf{Model-Free vs. Model-Based Approaches.}
Model-free approaches, such as the RMA \cite{RMA} and DATT \cite{DATT} frameworks, have been shown to be effective in learning adaptive control policies directly from interaction data. However, they often suffer from poor sample efficiency and limited flexibility at test time. In particular, incorporating new objectives or constraints—such as collision avoidance—typically requires retraining the policy. In contrast, model-based approaches offer greater flexibility by explicitly modeling system dynamics, allowing new cost functions or constraints to be incorporated at test time without retraining the entire framework. This makes model-based methods more suitable for scenarios requiring adaptability not only to changing environments but also to evolving task specifications.

\begin{table}[h]
\caption{Related work on adaptive strategies in robotic control.}
\label{tab:related-work}
\resizebox{\columnwidth}{!}{%

\begin{tabular}{lllllll}
\hline
\textbf{Paper} & \textbf{Method} & \textbf{NODE} & \textbf{Adaptation} & \textbf{Privileged Info.} & \textbf{Nominal-model-free} & \textbf{Real Test} \\ \hline
\cite{RMA} & \multirow{2}{*}{Model-free} & \textcolor{red}{\xmark} & \textcolor{blue}{\checkmark} & \textcolor{blue}{\checkmark} & \textcolor{blue}{\checkmark} & \textcolor{blue}{\checkmark} \\
\cite{DATT} &  & \textcolor{red}{\xmark} & \textcolor{blue}{\checkmark} & \textcolor{blue}{\checkmark} & \textcolor{red}{\xmark} & \textcolor{blue}{\checkmark} \\ \hline
\cite{nagabandi2018learning, 10341771} & \multirow{6}{*}{Model-based} & \textcolor{red}{\xmark} & \textcolor{blue}{\checkmark} & \textcolor{red}{\xmark} & \textcolor{blue}{\checkmark} & \textcolor{blue}{\checkmark} \\
\cite{xie2023hierarchicalmetalearningbasedadaptivecontroller, O_Connell_2022, levy2025metalearningonlinedynamicsmodel} &  & \textcolor{red}{\xmark} & \textcolor{blue}{\checkmark} & \textcolor{blue}{\checkmark} & \textcolor{red}{\xmark} & \textcolor{blue}{\checkmark} \\
\cite{meta_baseline} &  & \textcolor{red}{\xmark} & \textcolor{blue}{\checkmark} & \textcolor{blue}{\checkmark} & \textcolor{blue}{\checkmark} & \textcolor{blue}{\checkmark} \\
\cite{pmlr-v229-kasaei23a} &  & \textcolor{blue}{\checkmark} & \textcolor{red}{\xmark} & NA & NA & \textcolor{blue}{\checkmark} \\
\cite{pmlr-v205-jiahao23a} &  & \textcolor{blue}{\checkmark} & \textcolor{blue}{\checkmark} & \textcolor{red}{\xmark} & \textcolor{red}{\xmark} & \textcolor{blue}{\checkmark} \\
\cite{duong2022adaptivecontrolse3hamiltonian} &  & \textcolor{blue}{\checkmark} & \textcolor{blue}{\checkmark} & \textcolor{blue}{\checkmark} & \textcolor{red}{\xmark} & \textcolor{red}{\xmark} \\ \hline
\textbf{Ours} & \textbf{Model-based} & \textcolor{blue}{\checkmark} & \textcolor{blue}{\checkmark} & \textcolor{blue}{\checkmark} & \textcolor{blue}{\checkmark} & \textcolor{blue}{\checkmark} \\ \hline
\end{tabular}%
}
\end{table} 


\section{PROBLEM STATEMENT}
\begin{figure*}
  \centering
\includegraphics[width=0.75\textwidth]{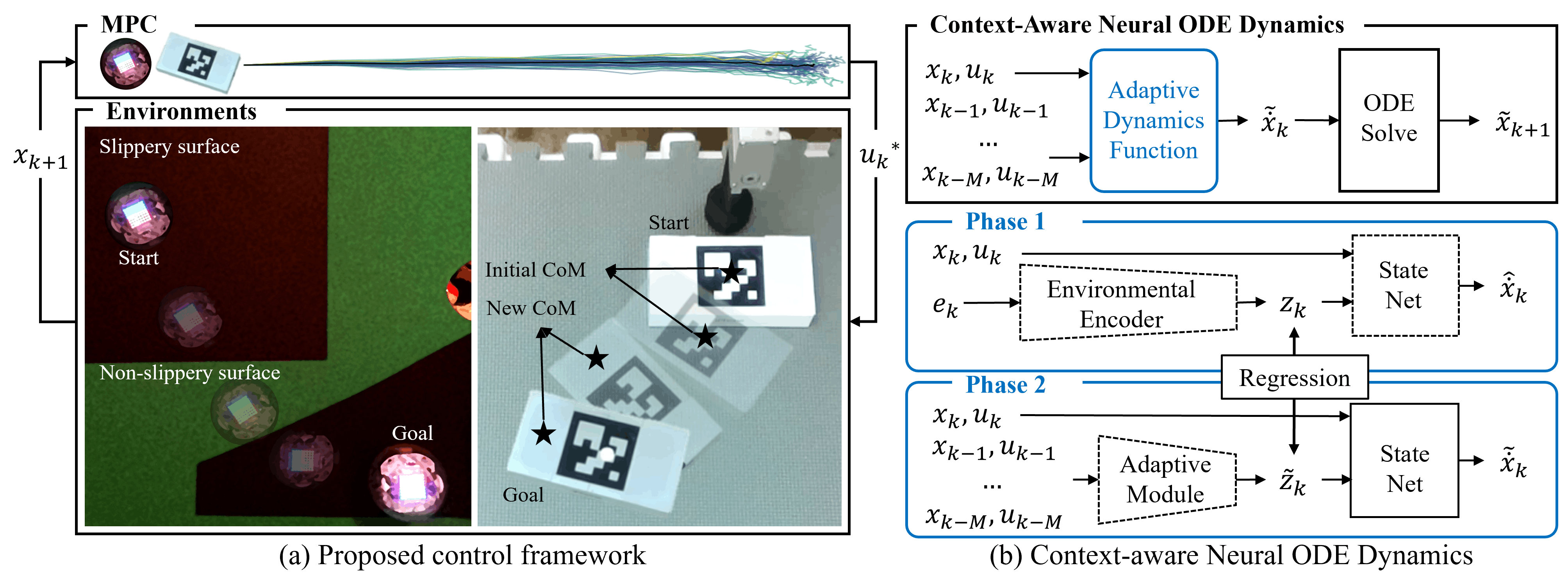}
  \caption{(a) Proposed control framework for robotic control, where MPC is adopted to determine optimal control actions by predicting future trajectories with our proposed dynamics model.
(b) Structure of the context-aware dynamics: State Net models the derivatives of states evolution, the environmental encoder processes privileged information, and the adaptive module reconstructs a latent environmental vector from historical state-action data by regressing on the corresponding latent vector from Phase 1. State prediction is obtained through numerical integration of the dynamics function. Models with trainable weights are indicated with dashed lines.}
  \label{fig:framework}
\end{figure*}

We formulate robot goal reaching and path tracking as a discrete-time MPC problem, which optimizes a sequence of future control actions over a finite time horizon. We consider a dynamical system governed by

\begin{align}
\label{eq:dynamics_derivative}
\dot{x}_{k} &= f(x_k, u_k, e_k), \nonumber \\
x_{k+1} &= x_{k} + \int_{t_{k}}^{t_{k+1}} f(x_k, u_k, e_k)\, dt
\end{align}

The state of the system at time step $k$ is denoted by $x_k \in \mathbb{R}^n$, the control input by $u_k \in \mathbb{R}^m$, and environmental factors (such as ground textures, wind fields, and payloads), which influence the system dynamics, by $e_k \in \mathbb{R}^l$. The function $f$ represents the system dynamics, modeling $\dot{x_{k}}$. A numerical integrator, such as the \textit{Euler} method or a higher-order \textit{Runge–Kutta} solver, can then be used to obtain $x_{k+1}$. Given a prediction horizon of length \( H \), the task for this system becomes an optimization problem that can be solved using MPC at each time step \( k \):
\begin{equation}
\label{eq:mpc-cost}
\begin{aligned}
\min_{u_k, \dots, u_{k+H-1}} \quad & \sum_{i=k}^{k+H-1} \ell(x_i, u_i) + \ell_f(x_{k+H}) \\
\text{s.t.} \quad 
& x_{i+1} = \text{ODESolve}(x_i, u_i, f, t_i, t_{i+1}), \\
& \hspace{2em} \forall i = k, \dots, k+H{-}1, \\
& u_i \in \mathcal{U}, \quad \forall i = k, \dots, k+H{-}1, \\
& x_0 \in \mathcal{X}_0
\end{aligned}
\end{equation}
Here, the stage cost and the terminal cost are denoted by $\ell$ and $\ell_f$, respectively. The input constraint set is denoted by $\mathcal{U}$, and the initial constraint set, designed to ensure that the robot starts from the designated state, is denoted by $\mathcal{X}_0$. After solving the problem~\eqref{eq:mpc-cost}, only the first control input $u_k^\star$ of the optimal sequence is applied to the system. In the next time step, the horizon is shifted forward, and the optimization is repeated with updated state information. MPC can handle a certain degree of uncertainty or disturbances by solving Equation (\ref{eq:mpc-cost}) online during execution. However, it fails to perform well when the system deviates significantly from the reference dynamics model. Therefore, a dynamics model that can mitigate model mismatch is crucial for maintaining control performance.

\section{Context-Aware Neural ODE Dynamics}

\subsection{Two-Phase Framework}
\label{subsec:two-phase}

In this section, we discuss how to learn the dynamics mapping from the current state $x_k$ and current action $u_k$ to the next state $x_{k+1}$ conditioned on historical data. The objective of learning the context-aware dynamics is to capture environments based solely on historical data, so that the dynamics function can be adjusted according to the inferred environment. While most model-based RL methods learn the mapping in an end-to-end manner, we decompose dynamics learning into two phases as shown in  Fig.~\ref{fig:framework}(b).

In Phase 1, we use privileged information $e_k$, which contains direct and complete environmental information, to facilitate learning the state evolution in response to control actions. However, since $e_k$ may not be measurable during deployment, we apply an \textit{environmental encoder} to first transform $e_k$ into a latent vector $z_k$. We then feed $z_k$ into \textit{State Net}, which is part of the dynamics function $f$, to obtain $\hat{\dot{x_k}}$, and use a numerical integrator to compute the next state $x_{k+1}$. This process models trajectories as solutions to a continuous-time differential equation, inherently producing smooth and physically consistent outputs. After completing end-to-end training in Phase 1, Phase 2 addresses the original mapping problem by learning to infer $z_k$ from historical data. Since Phase 1 is dedicated to modeling the temporal evolution of system states, we freeze the weights of the learned State Net in Phase 2. We then regress the historical data on their corresponding $z_k$ and obtain $\Tilde{z_k}$. The process can be expressed as: 
\vspace*{-0.2em}
\begin{equation}
\label{eq:measurement}
z_k = g(e_k),\;
\tilde{z}_k = h(\{(x_i, u_i)\}_{i=k-M}^{k-1}),\;
L = L_2(z_k, \tilde{z}_k)
\end{equation}

State-action history over a horizon of length $M$ is denoted by $\{(x_i, u_i)\}_{i=k-M}^{k-1}$, and regression loss is denoted by $L$. The environmental encoder, denoted by $g$, encodes $e_k$ into $z_k$. The adaptive module, denoted by $h$, reconstructs key information pertaining to the current environment by encoding the state-action history into $\tilde{z}_k$ and regressing it on $z_k$ using an MSE loss. We treat the environment in latent space because it is easier to align the domain of $\{(x_i, u_i)\}_{i=k-M}^{k-1}$ and its corresponding $e_k$ in another lower-dimensional space, as they carry distinct physical meanings and have significantly different dimensionality. 

\subsection{Training Details}
\label{subsec:training_detail}
To implement MPC, we use the Model Predictive Path Integral (MPPI) framework \cite{williams2017model} to obtain optimal control actions. To improve long-horizon prediction accuracy, we apply curriculum learning to train NODE with gradually increasing prediction horizons, from 1-step to H-step alignment. This mitigates the gradient explosion and vanishing issues that commonly arise when training on long sequences or fine-tuning an existing model. In addition, for the adaptive module design, using a 1D convolutional neural network to handle the high dimensionality of historical data shows benefits in quadrotor experiments by improving latent vector reconstruction through the extraction of local temporal patterns. The sliding filters capture environmental changes regardless of their position in the sequence, which is useful in robotic motion, where changes in velocity or motion direction may occur at arbitrary points within a time sequence. Since each element in the state and action vectors represents a distinct physical quantity, treating them as separate channels further enhances feature extraction.

To improve performance across environments, we incorporate online fine-tuning of the learned dynamics model. As new observations $\{(x_k, u_k, x_{k+1})\}$ become available during execution, we update the parameters in the dynamics model on the fly to reduce prediction errors. To avoid catastrophic forgetting as well as to balance exploration and exploitation, we use experience replay buffers to record all the observations for online learning and enable a robot to select between a random action and $u_k^*$. This continual learning process allows the model to refine its predictions and adapt to distributional shifts, especially for unseen historical data. Notably, we use the same training and testing time step for NODE throughout the experiments.

\section{Quadrotor Navigation in Simulation}

\begin{table*}[h]
\centering
\vspace{2mm}
\caption{Baselines and Ablations. CaDM refers to Context-aware Dynamics Model \cite{context_aware}. PPO refers to Proximal Policy Optimization \cite{schulman2017proximal}.}
\label{tab:baselines}
\begin{adjustbox}{max width=\textwidth} 
\begin{tabular}{lll}
\hline
\textbf{Type}    & \textbf{Method}           & \textbf{Description}                                                                                                                                                    \\ \hline
\multirow{4}{*}{Baselines} & Meta-learning dynamics        & We adopt the
concept of the meta-learning–based approach in \cite{meta_baseline} to design a context encoder using variational inference. \\
                           & CaDM & End-to-end adaptive dynamics with forward and backward losses to extract environmental factors \cite{context_aware}.                             \\
                           & RMA                           & Building on RMA \cite{RMA}, we enable trajectory tracking by embedding desired future positions into the PPO policy \cite{schulman2017proximal}.                           \\
                           & DATT                          & Model-free method using L1 adaptation to derive the environmental factors, trained with PPO \cite{DATT}.                         \\ \hline
\multirow{3}{*}{Ablation}  & MLP (Phase 1)                 & MLP-based dynamics model using privileged information, trained autoregressively with L2 loss over a fixed horizon.                                                             \\
                           & MLP (Phase 2)                 & MLP-based dynamics trained under the same two-phase framework using historical information.                                                                             \\
                           & Fixed NODE                    & NODE-based model without online adaptation, relying only on initial environmental factors during inference.                                                             \\ \hline
\multirow{2}{*}{Proposed}  & Ours (Phase 1)                & NODE-based model trained with privileged information as an upper bound, solved via \textit{Forward Euler} integration.                                                           \\
                           & Ours (Phase 2)                & Context-aware NODE model with an encoder that reconstructs $z_k$ from historical state-action data.                                                                     \\ \hline
\end{tabular}
\end{adjustbox}
\end{table*}

\textbf{Data Collection.} We collect datasets using simulators by randomly sampling initial states and applying random actions. Each trajectory consists of 50 state–action pairs. If the dynamics model trained on the dataset fails to accurately capture the true system behavior, the online dynamics learning approach discussed in Section~\ref{subsec:training_detail} becomes essential, as it enables the robot to explore and cover critical portions of the state–action space required to accomplish the task.
\begin{figure*}[tb]
  \centering \includegraphics[width=\textwidth]{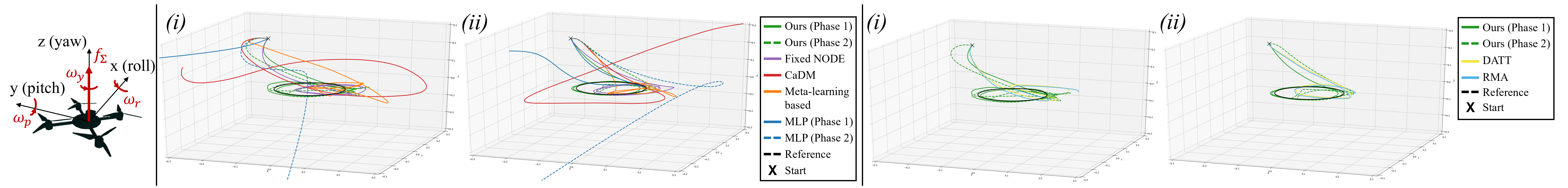} 
  \caption{Path tracking trajectories for the quadrotor under \textit{(i)} time-varying winds, and \textit{(ii)} spatially dissipating wind with random time-varying noise.}
\label{fig:path_tracking_traj}
\end{figure*}


\begin{table}[tb]
\caption{Performance of the quadrotor under \textit{(i)} time-varying winds, and \textit{(ii)} spatially dissipating wind with random time-varying noise. RMSE ($\mathrm{m}$).}
\label{tab:drone_env_all}
\centering
\resizebox{\linewidth}{!}{%
\begin{tabular}{lllll}
\hline
\textbf{}                   & \multicolumn{2}{c}{\textit{\textbf{i}}}                                 & \multicolumn{2}{c}{\textit{\textbf{ii}}}                                \\ \cline{2-5} 
                            & Goal reaching                      & Path tracking                      & Goal reaching                      & Path tracking                      \\ \hline
MLP(Phase 1)                & $>$ 0.2                            & $>$ 0.2                            & $>$ 0.2                            & $>$ 0.2                            \\
MLP(Phase 2)                & $>$ 0.2                            & $>$ 0.2                            & $>$ 0.2                            & $>$ 0.2                            \\
CaDM                        & $>$ 0.2                            & $>$ 0.2                            & $>$ 0.2                            & $>$ 0.2                            \\
Meta-learning based         & 0.092±0.023                        & 0.143±0.074                        & 0.099±0.021                        & 0.119±0.066                        \\
Fixed NODE                  & 0.066±0.011                        & 0.055±0.019                        & 0.065±0.008                        & 0.049±0.013                        \\
{RMA}  & {0.107±0.012} & {0.099±0.014} & {0.065±0.015} & {0.062±0.012} \\
{DATT} & {0.083±0.016} & {0.070±0.013} & {0.051±0.008} & {0.048±0.009} \\ \hline
Ours(Phase 1)            & \textbf{0.010±0.004}               & \textbf{0.036±0.020}               & \textbf{0.012±0.007}               & 0.034±0.012                        \\
Ours(Phase 2)            & 0.031±0.016                        & 0.049±0.026                        & 0.022±0.014                        & \textbf{0.030±0.018}               \\ \hline
\end{tabular}}
\end{table}

\textbf{Setup.} We use the quadrotor dynamics as the governing equation in our simulator \cite{DATT}. The state of the dynamics model is defined as $[p, v, q]^T$, where $p$, $v$ and $q$ are the 3D position, velocity, and quaternion. As shown in Fig. \ref{fig:path_tracking_traj}, control action is defined as $[f_{\Sigma}, \omega]^T$, where $f_{\Sigma}$ denotes the desired thrust and $\omega$ denotes the desired angular velocity in roll, pitch and yaw directions $[\omega_r, \omega_p, \omega_y]$. While the MPC determines high-level thrust and angular velocity commands, the low-level Proportional-Integral (PI) controller computes the necessary body torques; an inverse actuation matrix then maps this combined thrust and torque into individual motor speeds. To model a real-world quadrotor, we transform the continuous dynamics model into a discrete one with the sampling time set as $0.02\, \mathrm{s}$. Environmental variations are different wind fields acting on the quadrotor system. The wind field is modeled as a disturbance force along the x-direction. During data collection, we sample the disturbance force in the range of \([-1,1]\,\mathrm{N}\), where \(\mathrm{N}\) denotes Newtons, and collect trajectories under each wind field.

\begin{figure}
  \centering
  \includegraphics[width=0.9\columnwidth]{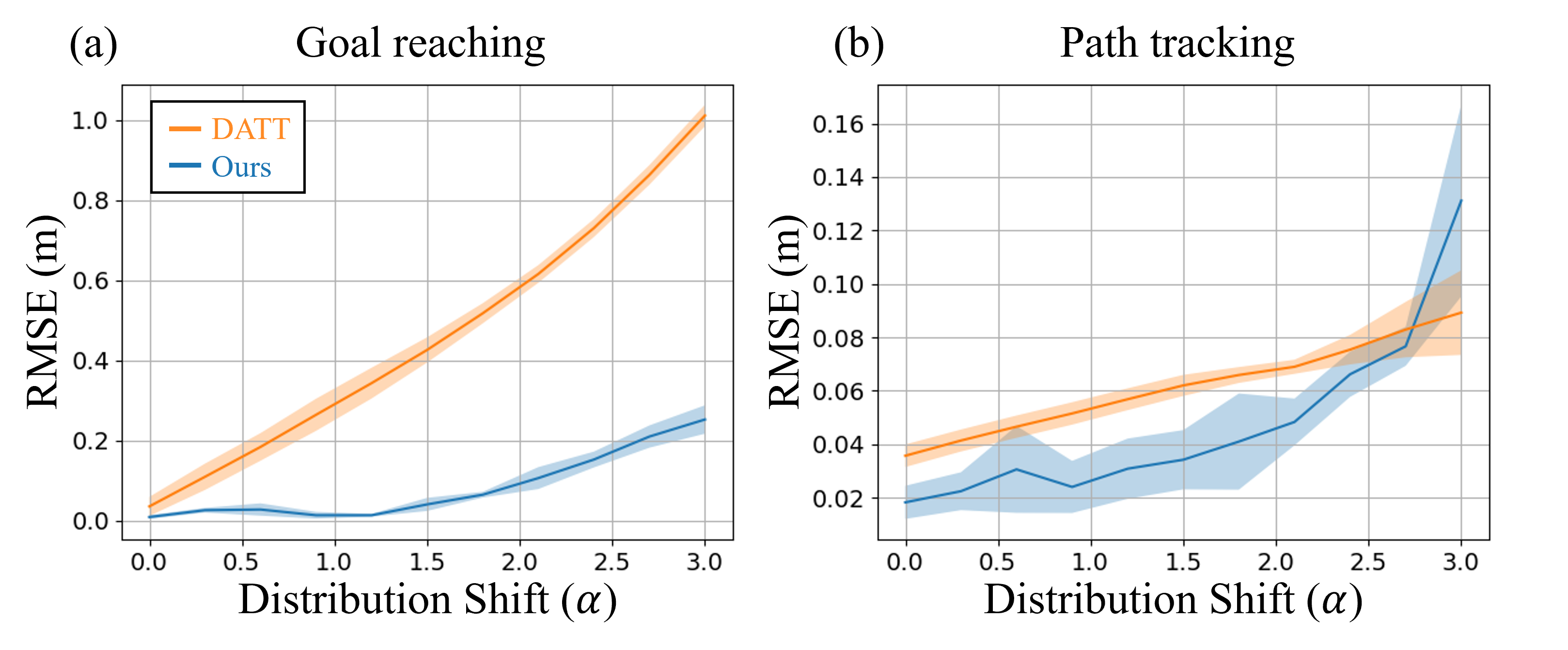}
  \caption{Performance degradation under distribution shift. (a) Goal-reaching and (b) path-tracking errors are plotted as functions of the distribution shift parameter $\alpha$. The training regime corresponds to $\alpha \in [0,1]$, while $\alpha > 1$ represents out-of-range conditions with increasing wind disturbance.}
  \label{fig:generalization}
\end{figure}

\textbf{Target tasks.} In the goal reaching and hovering task, the domain is a box with half-size $0.2\, \mathrm{m}$. The quadrotor starts at each vertex of the box, and the goal is located at the center. The objective is to control the quadrotor to reach and hover at the goal. We allow each controller $5\, \mathrm{s}$ to execute actions, then calculate the average position error (RMSE) between the quadrotor position and goal over the final second. In the path tracking task, the quadrotor begins at the same positions as in the goal reaching tasks and follows a circular path, evaluated with the same RMSE metric. Test environments include two wind conditions, spanning both within and beyond the training range.  $(i)$ time-varying winds fluctuating sinusoidally around nominal force of $1.5\, \mathrm{N}$ with the amplitude of $0.5\, \mathrm{N}$, updated every 10 control steps; and $(ii)$ a spatially dissipating wind field generated by a fan-like source located at $(-0.2,\,0,\,0)$, producing airflow in x direction. The wind magnitude decays with distance in both lateral and longitudinal directions from the source, with a peak force of $2\, \mathrm{N}$. The disturbance is further perturbed by additive Gaussian noise whose variance scales with the local wind magnitude, yielding a random time-varying component resembling Brownian motion \cite{DATT}.

The MPC cost function for a quadrotor performing a goal reaching and hovering or path tracking task is defined as $J = \sum_{k=0}^{H} \left( 
    w_p \left\| {p}_k - {p}_k^{\text{ref}} \right\|^2 
    + w_q \left( 1 - \left( {q}_k^\top {q}_k^{\text{ref}} \right)^2 \right)
\right)$, where ${p}_k^{\text{ref}}$ is the reference (goal or trajectory) position at time $k$, the reference quaternion at time $k$ is denoted by ${q}_k^{\text{ref}}$, and the weighting factors for the position and orientation errors are denoted by $w_p$ and $w_q$, respectively. 


\textbf{MPC performance.} We compare our method against the baseline and ablation methods listed in Table \ref{tab:baselines}. Table~\ref{tab:drone_env_all} shows the control performance. We also apply online dynamics learning in a constant wind field, as described in Section~\ref{subsec:training_detail}, to fine-tune the dynamics model. From the results, we observe that our approach successfully completes each navigation task under environmental conditions that vary across space and time, achieving lower tracking errors than the model-based baselines. While there is a slight performance drop from Phase 1 to Phase 2, the adaptive module in Phase 2 still reconstructs environmental factors and achieves adaptation. The quadrotor trajectories in the path tracking task (Fig.~\ref{fig:path_tracking_traj}) show that the meta-learning-based and fixed-dynamics models perform best among the baselines, successfully guiding the quadrotor near the target trajectory. However, these top baselines still cannot accurately track the reference, whereas our method achieves the best overall tracking performance.
{Table~\ref{tab:drone_env_all} also compares model-free methods (RMA and DATT) with our model-based approach. These two categories use fundamentally different training pipelines: PPO, a model-free and on-policy method, trains the policy directly from data sampled by the current policy, whereas our approach first collects offline data using random policies to train a dynamics model, and then fine-tunes the model using online data generated by the MPPI controller. Because of these differences, it is difficult to compare the methods using exactly the same training dataset. Nevertheless, we report the results in Table \ref{tab:drone_env_all} and show that our approach achieves comparable performance.}

The results in Fig.~\ref{fig:generalization} demonstrate the generalization capability of our model-based method compared to the model-free baseline, DATT, under increasing distribution shift in wind field with constant force. As the scaling factor $\alpha$ increases beyond the training range \([-1,1]\,\mathrm{N}\), both goal-reaching and path-tracking errors grow; however, our method exhibits significantly more graceful degradation, particularly in the goal-reaching task. While errors exceeding $0.2\,\mathrm{m}$ are considered failures, DATT fails at around $\alpha \approx 1$ for goal reaching, whereas our method remains robust up to approximately $\alpha \approx 3$. For path tracking, DATT achieves stronger performance—likely because it is explicitly trained for this task—yet both methods remain below the failure threshold even at $\alpha = 3$. This contrast highlights that our approach is not specialized for a single task but instead maintains consistent performance across different objectives. Moreover, the improved robustness can be attributed to the use of learned dynamics from transitions, whereas DATT directly learns a policy, making it more sensitive to distribution shifts beyond the training regime. Notably, our method is trained with $9M$ transition samples to learn the dynamics, whereas DATT requires $25M$ environment interactions to learn the policy, highlighting the better sampling efficiency of our approach. On the other hand, model-free methods offer significantly faster inference: PPO runs at roughly $1\, \mathrm{ms}$ per step—an order of magnitude faster than our $17\, \mathrm{ms}$—making them attractive for applications requiring very high control rates or limited onboard compute.
To further evaluate the effectiveness of online learning, we conduct experiments under wind disturbances exceeding $3\, \mathrm{N}$. After 8 hours of offline learning, the policy achieves a position RMSE of $0.1649 \pm 0.0221\,\mathrm{m}$ on path-tracking tasks. With an additional 2 hours of online finetuning, the RMSE is reduced to $0.0646 \pm 0.0153\,\mathrm{m}$.

\section{Real-World Experiments}

\subsection{Setup}

\begin{table*}[tb]
\vspace{2mm}
\centering
\caption{Control problem definition for real-world platforms: mobile robot and planar pushing.}
\label{tab:control_problem}
\resizebox{\textwidth}{!}{%
\begin{tabular}{|l|p{10cm}|p{10cm}|}
\hline
\textbf{} & \textbf{Mobile Robot Platform} & \textbf{Planar Pushing Platform} \\
\hline\hline

\textbf{State} & 
$\mathbf{s} = [x, y, dx, dy, \dot{x}, \dot{y}]$: 2D position $(x,y)$, heading direction $(dx, dy)$ as cosine/sine, linear velocity $(\dot{x}, \dot{y})$ & 
$\mathbf{s} = [x, y, \theta]$: object position $(x,y)$, object orientation $\theta$ \\
\hline

\textbf{Action} & 
$\mathbf{a} = [u_{\text{forward}}, u_{\text{turn}}]$: forward velocity command $u_{\text{forward}}$, steering angle command $u_{\text{turn}}$ & 
$\mathbf{a} = [u_{len}, u_{angle}, u_{loc}]$: pushing length $u_{len}$, pushing direction $u_{angle}$ with zero defined as the direction normal to the pushing surface, pushing location $u_{loc}$ with zero defined as the center of the pushing surface \\
\hline

\textbf{State Estimation} & 
\textit{Position:} Blob detection on two colored LED patches; robot center = average of patch centers. \textit{Heading:} $(dx, dy)$ the vector from one patch center to another. \textit{Velocity:} Finite-difference estimate of positions over two frames. & 
\textit{Position and Orientation:} Pose estimation of ArUco marker. \\
\hline

\textbf{Control Timing} & 
Camera: $60\,\mathrm{fps}$ ($\Delta t = 0.0167\, \mathrm{s}$); Control: $15\,\mathrm{Hz}$ ($4\times$ decimation); Inference time: $< 0.01\,\mathrm{s}$ (RTX 3090 GPU) & 
Camera: $8\,\mathrm{fps}$ ($\Delta t = 0.125\, \mathrm{s}$); Control: N/A (controller will wait for the action to complete before starting a new action); Inference time: $< 0.01\,\mathrm{s}$ (RTX 3090 GPU) \\
\hline

\textbf{Environment} & 
Pool table surface with randomized friction layouts (paper patches = low friction; table surface = high friction) & 
A box with a randomly placed internal weight. CoM denotes the location of the internal weight. \\
\hline

\end{tabular}%
}
\end{table*}

\begin{figure}
  \centering
  \includegraphics[width=0.75\columnwidth]{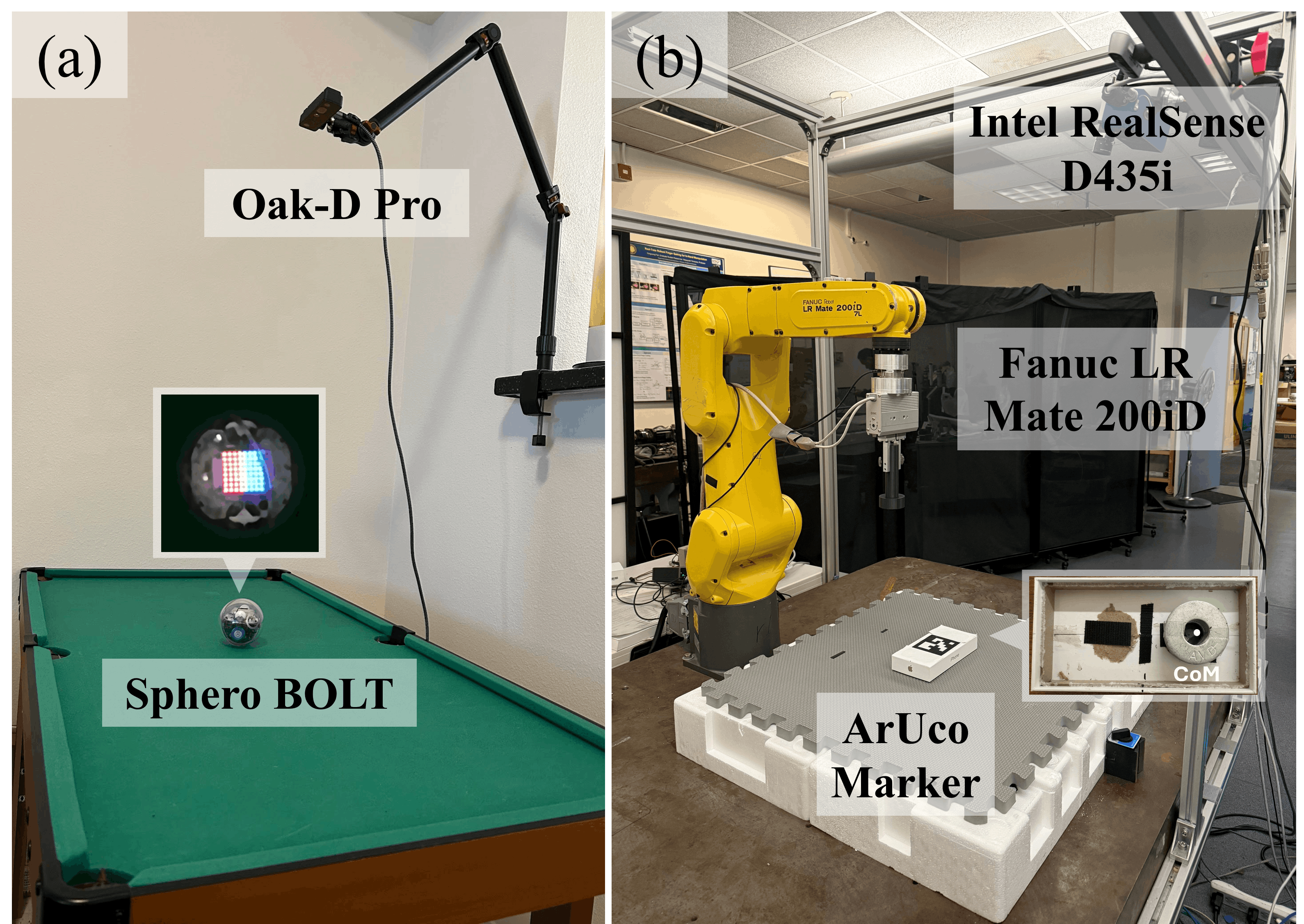} 
  \caption{ Real-world setups for (a) Sphero navigation and (b) manipulator pushing tasks.}
  \label{fig:real_world_setup}
\end{figure}

To demonstrate the real-world applicability of our algorithm, we evaluate the proposed framework on a mobile robot platform and a planar pushing platform, as shown in Fig.~\ref{fig:real_world_setup}. The detailed setup and control problem definition for both platforms can be found in Table \ref{tab:control_problem}.

\textbf{Mobile Robot Platform} \quad A Sphero BOLT robot is operated on a pool table and a Luxonis Oak-D Pro camera is mounted above the table to detect robot state. This platform evaluates adaptability to different surface textures.

\textbf{Planar Pushing Platform} \quad A Fanuc LR Mate 200iD manipulator with a cylindrical pusher is used to push a box toward a target, while an Intel RealSense D435i camera tracks its pose via an ArUco marker. The robot performs a slow, quasi-static push from the box’s long edge, and varying the internal weight shifts the CoM, altering the dynamics in ways that are not observable from the camera, making this a challenging testbed for CoM variation.

\subsection{Data Collection}

\textbf{Mobile Robot Platform} \quad Since the low-level controller in the Sphero BOLT robot is embedded in the robot's computational board (unknown to users) and the uncertainty of the robot's dynamics is complicated, it is hard to build a simulation environment that has small sim-to-real gap to the real-world platform. Therefore, we choose to collect data directly on the real-world platform by commanding the robot with random actions from a randomly placed location on the pool table. In sum, we collect around 2,000 pairs of (state, action, next state) for each texture. For privileged information in Phase 1 training, we assign relative friction coefficients to each texture. The friction coefficient is measured by traveling distance and velocity profile along the path when the robot is applied with the same action on each texture. Since we do not collect data for situations where the robot transitions between paper and pool table textures, we rely on the learned dynamics to generalize to these mixed-friction scenarios.

\textbf{Planar Pushing Platform} \quad Due to the quasi-static nature of the pushing action and high precision of the manipulator, we can build a simulator with minimal sim-to-real gap by doing system identification on real-world pushing trajectories. We build the simulation in MuJoCo and collect 116,000 pairs of (state, action, next state) for each of three different weight placements: centered, offset by $+1.8\,\mathrm{cm}$, and offset by $+3.6\,\mathrm{cm}$. After the Phase 1 model is trained, we combine it with MPC to collect rollout data for training the adaptive module in Phase 2.

\begin{table*}[tb]
\vspace{2mm}
\caption{Performance of real-world mobile robot navigation tasks across different friction layouts.
RMSE ($\mathrm{cm}$).}
\label{tab:real_world_navigation}
\centering
\resizebox{0.75\textwidth}{!}{%
\begin{tabular}{lcccccc}
\hline
\multicolumn{1}{c}{} 
& \multicolumn{2}{c}{Goal reaching} 
& \multicolumn{2}{c}{Circle tracking}
& \multicolumn{2}{c}{Triangle tracking} \\ \cline{2-7}
                     & Layout 1 & Layout 2 
                     & Layout 1 & Layout 2 
                     & Layout 1 & Layout 2 \\ \hline
Fixed NODE           & 2.127 ± 1.233 & 2.725 ± 1.176 
                     & 4.567 ± 2.106 & 2.858 ± 0.744 
                     & 5.404 ± 0.193 & 5.124 ± 0.318 \\
Ours                 & \textbf{0.766 ± 0.303} & \textbf{0.372 ± 0.082} 
                     & \textbf{2.077 ± 0.340} & \textbf{2.417 ± 0.124} 
                     & \textbf{3.315 ± 0.027} & \textbf{2.681 ± 0.113} \\ \hline
\end{tabular}
}
\end{table*}

\begin{table*}[tb]
\caption{Performance of real-world planar pushing tasks with varying CoM locations.}
\label{tab:real_world_pushing}
\centering
\resizebox{\textwidth}{!}{%
\begin{tabular}{lccccccc}
\hline
           & \multicolumn{5}{c}{RMSE (cm)}                                                                                              & \multicolumn{1}{l}{\multirow{2}{*}{Computation time (ms)}} & \multicolumn{1}{l}{\multirow{2}{*}{\# of (state, action, next state)}} \\ \cline{2-6}
           & Training               & Moderate               & Extreme                & Time varying           & Disturbance       & \multicolumn{1}{l}{}                                     & \multicolumn{1}{l}{}                                                   \\ \hline
Fixed NODE & 4.781 ± 0.571          & 4.587 ± 0.516          & \textbf{2.285 ± 0.411} & 3.368 ± 0.643          & 4.174 ± 0.740          & 6.637 ±0.145                                             & \textbf{348K}                                                          \\
RMA        & 0.471 ± 0.120          & 1.265 ± 0.205          & 7.072 ± 0.175          & \textbf{0.491 ± 0.035} & 9.548 ± 1.932          & \textbf{0.398 ± 0.069}                                   & 600K                                                                   \\
Ours       & \textbf{0.447 ± 0.256} & \textbf{1.129 ± 0.302} & 2.408 ± 0.796          & 1.040 ± 0.314          & \textbf{1.713 ± 1.918} & 7.799 ± 0.236                                            & 355K                                                                   \\ \hline
\end{tabular}
}
\end{table*}

\begin{figure*}
  \centering
  \includegraphics[width=\textwidth]{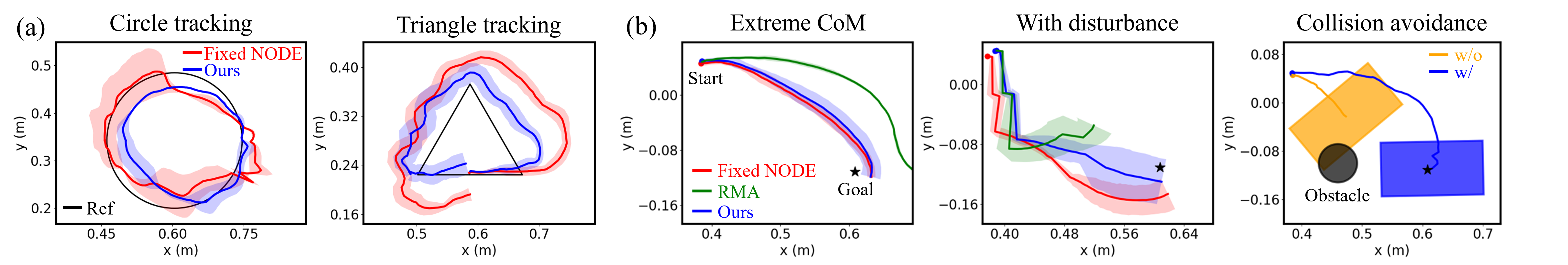} 
  \caption{Trajectories for (a) mobile robot navigation and (b) planar pushing tasks.}
  \label{fig:real_world_traj}
\end{figure*}
\subsection{Results on Mobile Robot Platform}

We evaluate the platform on two tasks: goal reaching and path tracking. In the goal reaching task, each layout is tested from a random initial position, with the table center serving as the goal. The robot must reach and hover at the goal, and performance is measured by the minimum distance between its trajectory and the goal. For the path tracking task, we test two reference tracks: a circular path and a triangular path that provides a more aggressive reference. Both reference paths begin from the same initial position. The robot is required to track the reference path, and performance is evaluated using the mean Euclidean distance between the trajectory and the nearest point on the path. We design two cost functions for the tasks. $J_1$ is for goal reaching and $J_2$ is for path tracking. 
The two definitions are expressed as $J_1=\sum_{k=0}^{H} \left[ 
    w_p  \left\| {p}_k - {p}^{\text{ref}}_k \right\|^2 +
    w_\theta  \left( \theta_k - \theta_k^{\text{pp}} \right)^2 
\right]$ and $J_2 = \sum_{k=0}^{H} \left[
    w_p  \left\| {p}_k - {p}^{\text{ref}}_k \right\|^2 +
    w_v  \left( v_k - v^{\text{ref}}_k \right)^2 +
    w_\theta  \left( \theta_k - \theta^{\text{ref}}_k \right)^2
\right]$, where \( p_k = [x_k, y_k]^T \) denotes the position, with reference \( p_k^{\text{ref}} \); \( v_k = \|[\dot{x}_k, \dot{y}_k]\|_2 \) denotes the velocity with reference \( v_k^{\text{ref}} \); and \( \theta_k^{\text{pp}} = \arctan2(y_{\text{goal}} - y_k, x_{\text{goal}} - x_k) \) denotes the pure-pursuit heading toward the goal with reference \( \theta_k^{\text{ref}} \). The positive scalar weights for position, heading, and velocity errors are denoted by \( w_p \), \( w_\theta \), and \( w_v \), respectively.

Table \ref{tab:real_world_navigation} reports results on two friction layouts, each averaged over three runs from similar start poses. Since the fixed NODE-based dynamics does not adapt to environment changes, it uses the friction coefficient at the start location and treats environmental changes as disturbances. The results show that our model achieves lower errors, indicating effective adaptation to spatially varying friction. It also yields a smaller standard deviation, demonstrating greater robustness and higher repeatability under real-world uncertainty. Our model is deployable on real-world systems and outperforms fixed NODE-based dynamics. Our model also handles surface boundary crossings more reliably, whereas the fixed NODE often gets stuck or loses track. The videos are available \href{https://youtu.be/PY0sNyF2rqE}{here}, and the trajectories for path tracking are shown in Fig.~\ref{fig:real_world_traj}.

We note that the reported control frequency of $15\,\mathrm{Hz}$ is not limited by the computational cost of the proposed method. In our system, the combined MPC and NODE forward simulation requires less than $0.01\,\mathrm{s}$ per control step, corresponding to a potential control rate exceeding $100\,\mathrm{Hz}$. The primary bottleneck arises from Bluetooth communication latency when transmitting commands to the robot. Therefore, the observed control frequency reflects system-level communication constraints rather than the computational efficiency of the MPC or NODE components.

\subsection{Results on Planar Pushing Platform}

The task is to command the manipulator to push the box towards the same goal position from the start pose. The performance is measured by the minimum distance between the rollout trajectory and the goal position. The cost design is the L2 norm of the distance to the goal. Table \ref{tab:real_world_pushing} shows the controller performance on different locations of the mass in the box. \textbf{Training} refers to the in-distribution CoM ($+3.6\,\mathrm{cm}$ offset) seen during training. \textbf{Moderate} refers to the configuration ($+6.0\,\mathrm{cm}$ offset) where the inner weight touches the wall of the box and is not seen in the training data. \textbf{Extreme} refers to an unseen configuration with a $-2.5\,\mathrm{cm}$ offset on the opposite half of the box. This setting is considered extreme because its dynamic behavior is completely reversed compared to the positive-offset configurations. \textbf{Time varying} refers to a dynamic configuration where the weight starts at the center and shifts to a random positive offset after 10 control steps. \textbf{Disturbance} refers to the standard $+3.6\,\mathrm{cm}$ offset configuration, but introduces external perturbations by applying a negative y-axis force to the box at control steps 1, 5, and 20. All metrics are obtained from three trials starting from similar initial poses. For the fixed NODE-based approach, we use the learned dynamics assuming the CoM is at the center of the box, which is a reasonable assumption when the interior of the box is not observable.

The results in Table~\ref{tab:real_world_pushing} highlight the benefit of incorporating adaptation into the learned dynamics model. Compared to the Fixed NODE baseline, our method consistently achieves lower RMSE in most scenarios, demonstrating that online adaptation improves performance under distribution shifts. In the extreme setting, Fixed NODE exhibits comparable performance to our method; this can be attributed to its assumption that the internal weight is centered, which happens to be close to the $-2.5\,\mathrm{cm}$ CoM configuration. In contrast, our adaptation module does not explicitly converge to this exact value, as $-2.5\,\mathrm{cm}$ lies far outside the training distribution. When compared with the model-free RMA approach, our method achieves comparable accuracy in the training and moderate regimes, while significantly outperforming RMA in more challenging scenarios. In particular, our method reduces error by over $60\%$ in the extreme case and over $80\%$ under disturbance, indicating improved robustness to severe distribution shifts. Notably, this performance gain is achieved with approximately $40\%$ fewer training samples, highlighting the superior sample efficiency of our model-based approach. It is important to note that the model-free method relies on carefully engineered reward functions to achieve the performance in Table \ref{tab:real_world_pushing}. In our implementation, the reward is defined as $r = -d + \alpha \Delta d + \beta e^{-\gamma d}$, where $d$ is the distance to the goal in world xy plane, $\Delta d$ denotes the step-wise progress, and $(\alpha, \beta, \gamma) = (5.0, 2.0, 50)$ are weighting coefficients. We observe that using a simpler reward such as $r = -d$ leads to a performance degradation of approximately $3\times$, underscoring the sensitivity of model-free methods to reward design. From the results, we observe that our model-based approach does not require a tedious cost design process to achieve strong goal-reaching accuracy. RMA performs slightly better in the time-varying setting, where the inner weight is randomly placed over time; the sampled weight locations in this experimental setup may be more favorable to RMA, as they are more likely to fall within its training distribution. Additionally, RMA benefits from over $10\times$ faster inference compared to our method. However, our framework offers greater flexibility through the online optimization process. Fig.~\ref{fig:real_world_traj} shows that we enable collision avoidance in real-world deployments without retraining by incorporating an exponential barrier function of the form $c_{\text{obs}} = \lambda e^{-\kappa d_{\text{obs}}}$, where $d_{\text{obs}}$ is the distance to the obstacle in world xy plane and $(\lambda, \kappa)$ are design parameters, whereas RMA would require additional training to achieve the same capability. The videos are available \href{https://youtu.be/PY0sNyF2rqE}{here}, and the trajectories are shown in Fig.~\ref{fig:real_world_traj}.

\section{Conclusion \& Future works}

In this paper, we propose a context-aware Neural ODE dynamics model that integrates NODE with a two-phase training process to reconstruct environmental factors and adjust state predictions accordingly. Combined with MPC, experimental results show that our method achieves superior performance compared to existing dynamics models on quadrotor simulation platforms subject to a range of wind fields. We also demonstrate the effectiveness of applying the framework to real-world systems, including a Sphero BOLT robot and a Fanuc manipulator. A limitation is that performance degrades when training and testing time steps are mismatched, even though larger time steps are desirable in MPC for improved efficiency; this could be addressed in future work by training a time-conditioned NODE.

\section*{Acknowledgment}

The work was supported by FANUC, the Finnish Center for Artificial Intelligence (FCAI), and CSC -- IT Center for Science.

\ifCLASSOPTIONcaptionsoff
  \newpage
\fi



%

\bibliographystyle{IEEEtran}
\bibliography{references}
%








\end{document}